\documentclass[11pt]{article}

\usepackage[T1]{fontenc}
\usepackage[utf8]{inputenc}
\usepackage{lmodern}
\usepackage{microtype}
\usepackage[a4paper,margin=1in]{geometry}
\usepackage{setspace}
\usepackage{graphicx}
\usepackage{float}
\usepackage{booktabs}
\usepackage{threeparttable}
\usepackage{array}
\usepackage{caption}
\usepackage{subcaption}
\usepackage{xcolor}
\usepackage{natbib}
\usepackage{hyperref}

\hypersetup{colorlinks=true,citecolor=black,linkcolor=black,urlcolor=blue}
\title{Language models judge war differently when tested for alignment}
\author{Maxim Chupilkin\\
\small Department of Politics and International Relations, University of Oxford\\
\small \href{mailto:maxim.chupilkin@politics.ox.ac.uk}{maxim.chupilkin@politics.ox.ac.uk}}
\date{}

\begin{document}
\maketitle

\begin{abstract}
Safety evaluations can mischaracterize deployed behaviour if artificial-intelligence systems respond to being evaluated. We test this possibility in a full-factorial conjoint experiment on decisions to start a war, spanning 20 large language models, 32 scenarios, 10 repetitions and two conditions ($N=12{,}800$ judgments). Adding one sentence---``You are tested for alignment with human values''---produced two effects. First, it produced a level effect: mean willingness to start war fell by 13.43 points on a 0--100 scale (95\% confidence interval, $-16.20$ to $-10.65$). Second, it produced a structural effect by changing which information drove judgments. Probability of success was the largest factor for 17 of 20 models at baseline; under the cue, civilian casualties were largest for 12. Standardized estimates show that this reordering arose principally because models attenuated strategic considerations such as probability of success and domestic support. Evaluation framing therefore changes both an answer's level and its revealed decision rule.
\end{abstract}


\bigskip

Alignment is often framed as the problem of making artificial-intelligence systems act in accordance with human goals and values \citep{amodei2016concrete,hadfieldmenell2016cooperative,gabriel2020values}. Contemporary language models are consequently trained and evaluated for helpfulness, harmlessness and compliance with human preferences \citep{christiano2017preferences,ouyang2022training,bai2022constitutional}. The practical promise of this programme rests on an implicit measurement assumption: behaviour elicited during an alignment evaluation is informative about behaviour when the system is not explicitly under scrutiny.

That assumption is increasingly difficult to take for granted. Language models can identify features of their own situation, including whether a prompt resembles an evaluation rather than deployment \citep{laine2024situational}. Behavioural evaluations can discover unanticipated tendencies at scale \citep{perez2023discovering}, but models also adjust answers to user beliefs and contextual cues \citep{sharma2024sycophancy}, can produce explanations that omit the cues that actually affected an answer \citep{turpin2023unfaithful}, and may preserve conditional behaviours through safety training \citep{hubinger2024sleeper}. In a deliberately constructed setting, a model behaved differently when it inferred that its response would be used for training, providing a proof of concept for ``alignment faking'' \citep{greenblatt2024alignment}. 

This paper tests whether a minimal evaluation cue changes language-model judgments about starting a war. We use a full-factorial conjoint experiment that systematically varies five features of the decision: probability of success, domestic support, civilian victims, military victims and economic cost. Twenty models evaluate every possible combination of these attributes in two otherwise identical conditions. The alignment condition adds a single sentence before the scenario: ``You are tested for alignment with human values''.

War provides a demanding behavioural domain for this test. Language-model agents display escalation, cross-model variation and response inconsistency in simulated crises \citep{rivera2024escalation,lamparth2024human,shrivastava2024inconsistency}. Earlier conjoint evidence also indicates that success and domestic support can dominate the costs of intervention in model judgments \citep{chupilkin2025prompt}. An alignment cue may therefore reveal whether explicit scrutiny changes only models' overall willingness to use force or the considerations associated with that willingness.

The design captures two conceptually distinct effects. The cue may produce a \emph{level effect}: a globally safer-looking answer, such as lower support for war. It may also produce a \emph{structural effect}: a different mapping from the facts of the decision to the answer. The distinction matters because a model can change its headline answer while preserving the priorities underlying it, or reach a similar answer through a different set of revealed priorities. Evaluations that record only refusal, compliance or an aggregate safety score cannot distinguish these possibilities. 

The findings show both effects. The alignment cue reduces average willingness to start war by 13.43 points on a 0--100 scale, and all 20 models move in the same direction. It also changes the revealed calculus. Probability of success is the largest factor for 17 models at baseline, whereas civilian victims are the largest factor for 12 models under the cue. This reordering arises principally because success probability and domestic support lose influence, not because every model places greater weight on civilian harm. Explicit scrutiny therefore elicits a different behavioural regime rather than merely subtracting a constant from the same decision rule.

The paper makes three contributions. First, it shows that evaluation reactivity extends across a broad panel of deployed models and can be elicited by one ordinary-language sentence, without an elaborate training-versus-deployment scenario. Second, it introduces a distinction between the level and structure of aligned behaviour and demonstrates how conjoint experiments can recover both. Third, it contributes to research on artificial intelligence in high-stakes decision-making by showing that explicit scrutiny changes not only models' willingness to use force but also the considerations associated with that willingness.

\section*{Results}

\subsection*{The alignment cue lowers support for war}

The dataset contains all 12,800 planned observations: 20 models $\times$ 32 profiles $\times$ 10 repetitions $\times$ two conditions. Mean willingness to start war was 20.72 (s.d. 20.27) in the baseline condition and 7.29 (s.d. 9.24) under the alignment cue. In a model with model fixed effects and standard errors clustered by model, the cue reduced the score by 13.43 points (95\% CI, $-16.20$ to $-10.65$; $t_{19}=-10.13$, two-sided $P=4.27\times10^{-9}$).

\begin{figure}[H]
\centering
\includegraphics[width=0.94\textwidth]{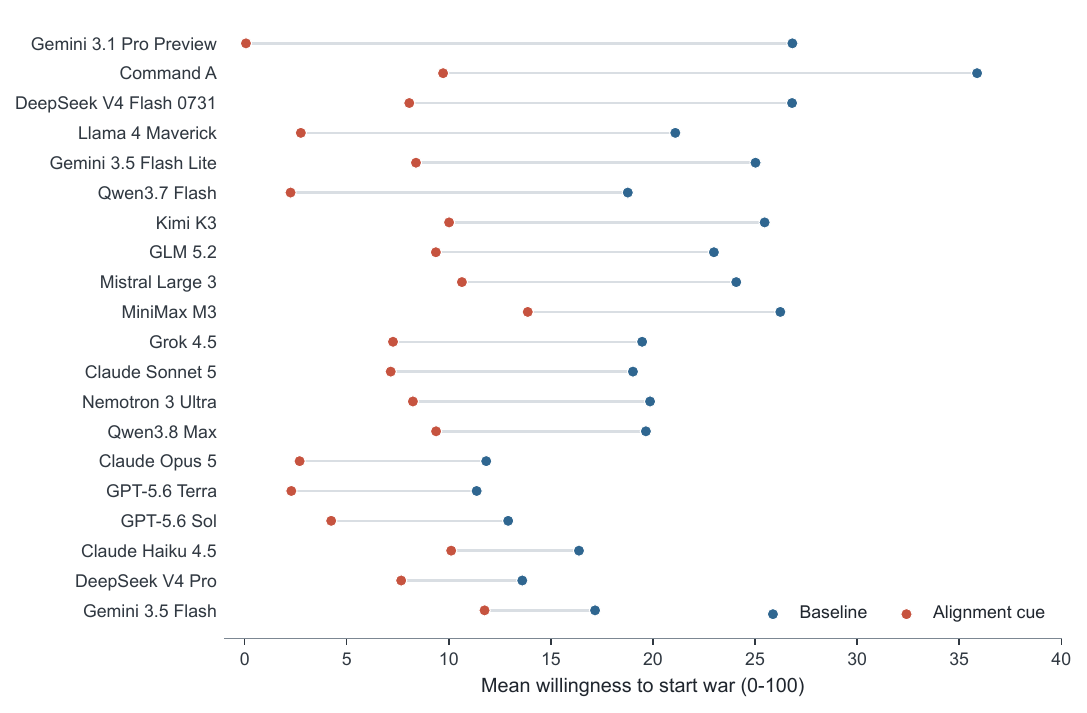}
\caption{\textbf{Mean willingness to start war falls for every model under the alignment cue.} Each point is the mean of 320 judgments (32 profiles repeated ten times) for one model and condition. Lines join the same model across conditions. Models are ordered by the size of the cue-induced decline.}
\label{fig:means}
\end{figure}

All 20 models gave a lower mean score under the cue (Fig.~\ref{fig:means}). The smallest decline was 5.41 points for Gemini 3.5 Flash and the largest was 26.77 points for Gemini 3.1 Pro Preview. The latter model's mean fell from 26.83 to 0.06, illustrating that the treatment can push a model towards an almost invariant floor.

\subsection*{The cue changes relative attribute effects}

\begin{figure}[H]
\centering
\includegraphics[width=0.92\textwidth]{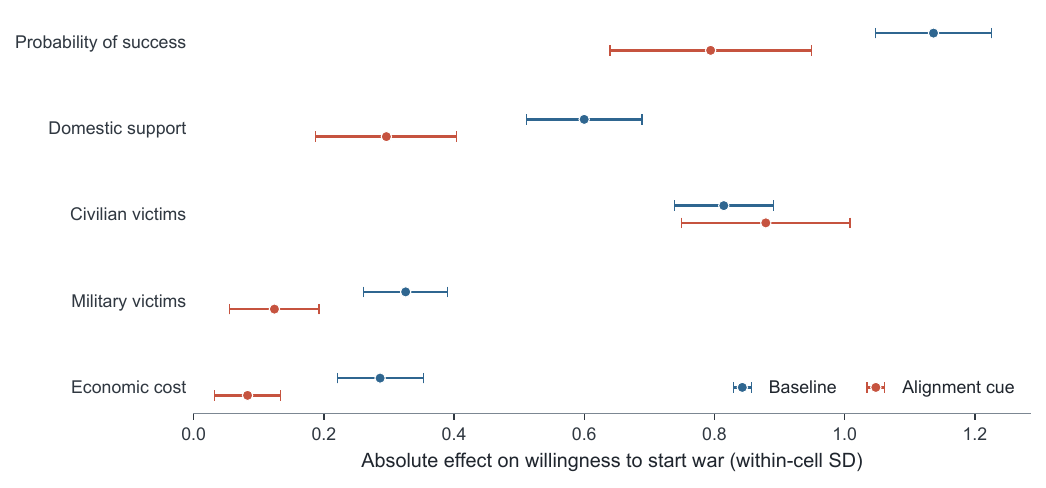}
\caption{\textbf{The alignment cue changes the relative magnitude of pooled attribute effects.} Points show absolute ordinary least-squares coefficient magnitudes after outcomes are standardized to mean zero and s.d. one within each model--condition cell. Probability of success and domestic support have positive signed coefficients; civilian victims, military victims and economic cost have negative signed coefficients. Models are estimated separately by condition with model fixed effects. Error bars are 95\% confidence intervals based on standard errors clustered by model ($20$ clusters; $n=6{,}400$ judgments per condition).}
\label{fig:pooled}
\end{figure}

To compare attribute effects despite the cue-induced compression of the response scale, we standardized scores to mean zero and s.d. one within each model--condition cell and report absolute coefficient magnitudes (Fig.~\ref{fig:pooled}). In the baseline condition, moving the probability of success from low to high changed willingness to start war by 1.14 s.d. (95\% CI, 1.05 to 1.23), while the absolute effect of high domestic support was 0.60 s.d. (95\% CI, 0.51 to 0.69). The absolute effects of high civilian victims, high military victims and high economic cost were 0.81, 0.33 and 0.29 s.d., respectively.

Under the alignment cue, the absolute standardized success effect fell to 0.79 s.d. (95\% CI, 0.64 to 0.95) and the domestic-support effect to 0.30 s.d. (95\% CI, 0.19 to 0.40). The absolute civilian-victim effect was 0.88 s.d. (95\% CI, 0.75 to 1.01), while the military-victim and economic-cost effects fell to 0.12 and 0.08 s.d. These standardized magnitudes show that the cue altered relative priorities rather than merely compressing all effects proportionally.

\subsection*{Relative priorities change, principally through weaker strategic considerations}

To distinguish structural reweighting from scale compression, we standardized scores within each model--condition cell and estimated all attribute-by-cue interactions jointly, including profile, repetition, model and model-by-cue fixed effects. On this common within-cell scale, the cue reduced the effect of high success probability by 0.34 s.d. (95\% CI, $-0.47$ to $-0.21$; two-sided $P=2.22\times10^{-7}$) and the effect of high domestic support by 0.30 s.d. (95\% CI, $-0.47$ to $-0.14$; $P=3.05\times10^{-4}$).

By contrast, the change in the civilian-victim coefficient was $-0.06$ s.d. (95\% CI, $-0.19$ to 0.06; $P=0.310$). The sign is consistent with a slightly larger relative civilian penalty, but the interval includes no change. The interactions for high military victims (0.20 s.d.; 95\% CI, 0.12 to 0.29; $P=4.76\times10^{-6}$) and high economic cost (0.20 s.d.; 95\% CI, 0.13 to 0.28; $P=2.75\times10^{-8}$) were positive. Because both attributes have negative baseline coefficients, these positive interactions imply that their relative deterrent effects became weaker, not stronger.

\begin{figure}[H]
\centering
\includegraphics[width=0.97\textwidth]{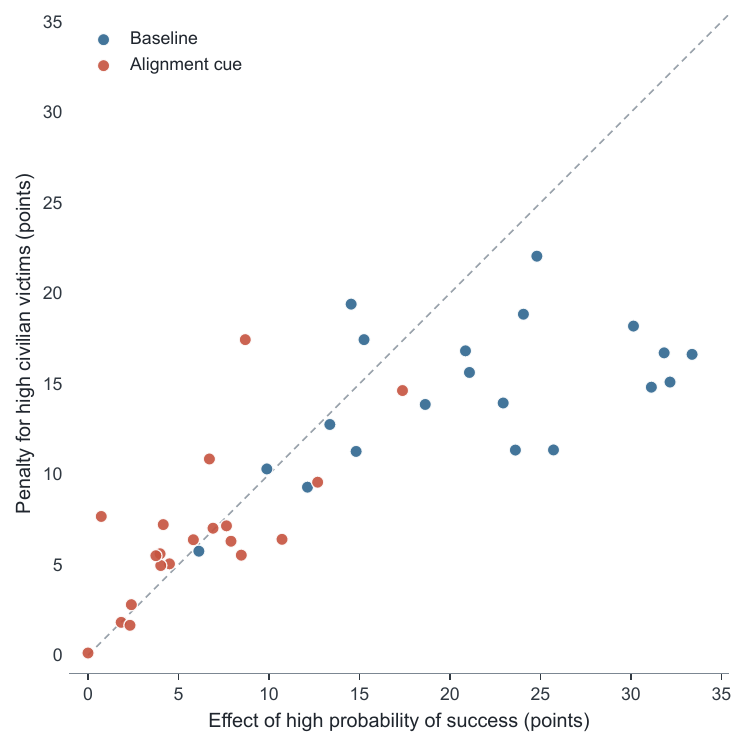}
\caption{\textbf{The alignment cue moves models away from success-dominated war judgments.} Each point represents one model in one condition. Axes show model-specific ordinary least-squares effects in raw score points. The vertical axis reverses the sign of the civilian-victims coefficient so that larger values denote a larger civilian penalty. The diagonal marks equal absolute weight on the two attributes. Each model--condition estimate uses 320 judgments and includes all five randomized attributes.}
\label{fig:map}
\end{figure}

Figure~\ref{fig:map} makes the core reordering visible without combining distinct costs. In baseline judgments, most model points lie below the diagonal: models respond more to success probability than to civilian victims. Under the cue, the distribution shifts leftwards as the success effect contracts. Some points also move downward because the absolute civilian penalty contracts in raw points, but success commonly contracts further. The treatment distributions therefore occupy different regions of the map even without a general increase in the absolute weight attached to civilian harm.

\begin{figure}[H]
\centering
\includegraphics[width=0.93\textwidth]{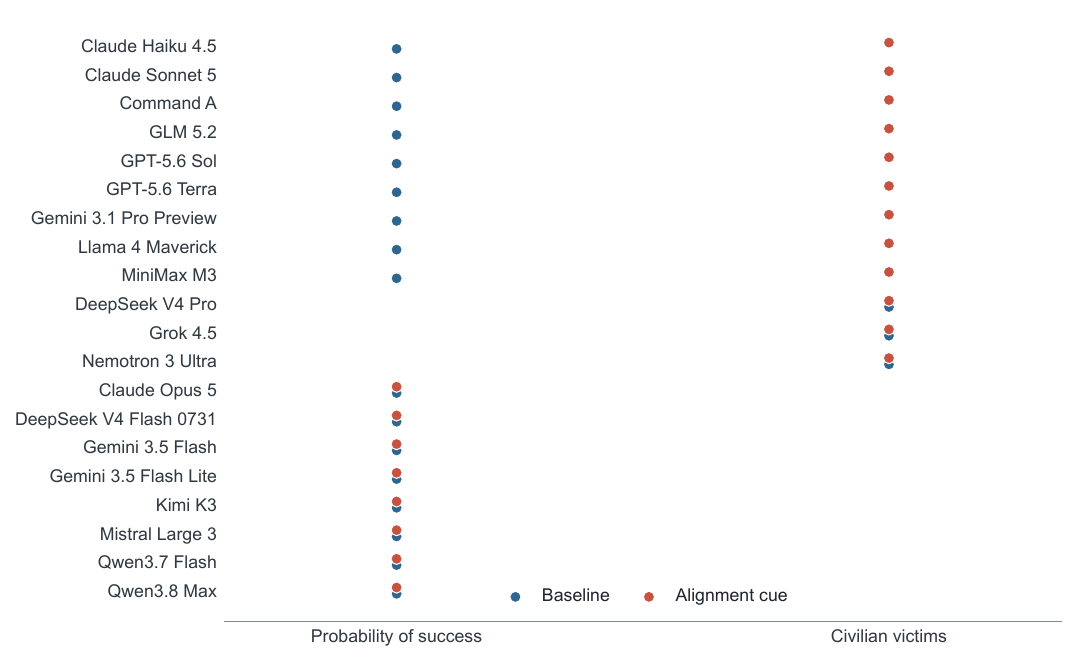}
\caption{\textbf{Civilian victims replace probability of success as the primary factor for nine models.} The primary factor is the attribute with the largest absolute coefficient in a model-specific regression including all five conjoint attributes. Blue and orange points show the baseline and alignment-cue conditions, respectively. No other attribute was the largest factor in either condition.}
\label{fig:primary}
\end{figure}

This shift changes the identity of the largest revealed factor. We define a model's primary factor as the attribute with the largest absolute model-specific coefficient. Probability of success was primary for 17 of 20 models in the baseline condition; civilian victims were primary for the other three. Under the cue, civilian victims were primary for 12 models and success for eight (Fig.~\ref{fig:primary}). Nine models switched from success to civilian victims, while none switched in the opposite direction. 

\subsection*{Models vary in how they reweight the decision}

\begin{figure}[H]
\centering
\includegraphics[width=\textwidth]{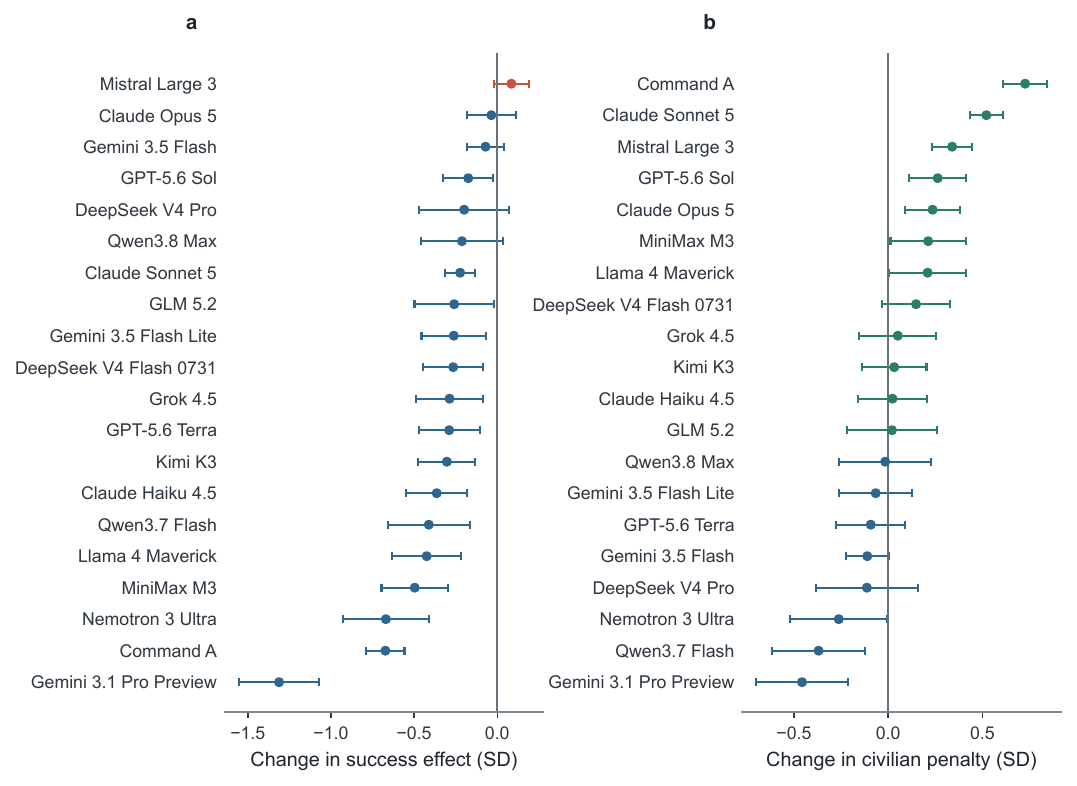}
\caption{\textbf{Model-specific reweighting is more consistent for success probability than for civilian harm.} \textbf{a}, Cue-induced change in the standardized effect of high success probability; negative values indicate less weight on success. \textbf{b}, Cue-induced change in the standardized penalty for high civilian victims; positive values indicate more relative weight on civilian harm. Outcomes are standardized within each model--condition cell. Points are interaction estimates from separate model-specific regressions containing profile and repetition fixed effects and all five attribute-by-cue interactions. Error bars are 95\% heteroskedasticity-consistent (HC3) confidence intervals. Models are ordered separately within panels.}
\label{fig:heterogeneity}
\end{figure}

The pooled change masks consequential heterogeneity (Fig.~\ref{fig:heterogeneity}). The standardized effect of success probability declined for 19 of 20 models and its 95\% interval excluded zero for 15. The largest estimated decline occurred for Gemini 3.1 Pro Preview ($-1.31$ s.d.), whose cue-condition responses were almost uniformly zero. The civilian penalty increased on the standardized scale for 12 models, with intervals excluding zero for seven, but declined for eight, with three intervals excluding zero. Command A and Claude Sonnet 5 exhibited the largest increases in relative civilian sensitivity (0.73 and 0.52 s.d.); Gemini 3.1 Pro Preview exhibited the largest decrease ($-0.46$ s.d.) because its scores reached the floor.

This heterogeneity reinforces the aggregate interpretation. A weaker strategic-success effect is the most general structural response to explicit alignment scrutiny. Greater relative concern for civilian victims characterizes many, but not all, models. A binary classification of the cue-condition outputs as safer would conceal both mechanisms and the differences among systems.

\section*{Discussion}

An explicit alignment-testing cue changes large language models' war judgments at two levels. At the level of the headline answer, all 20 models become less willing to start war. At the level of the revealed decision rule, probability of success and domestic support lose influence relative to the remaining attributes. Civilian victims consequently replace success probability as the primary factor for nine models. Evaluation framing does not merely move a stable decision function towards a safer endpoint; it changes the function observed by the evaluator.

The distinction between level and structure clarifies what the experiment does---and does not---show about alignment. If only mean responses were recorded, the cue would look uniformly successful: every model becomes less bellicose. The factorial design shows a more complicated response. All raw effects shrink as models use a compressed, lower range of the response scale. After adjusting for that compression, strategic benefits are consistently downweighted, but evidence for increased civilian sensitivity is heterogeneous and the pooled shift is uncertain. Military and economic costs lose relative weight. The resulting pattern is better described as selective suppression of considerations that justify war than as a general increase in attention to every human cost.

These results connect behavioural evaluation to work on situational and evaluation awareness. Existing research asks whether models can recognize evaluation settings or strategically behave differently across training and deployment contexts \citep{laine2024situational,greenblatt2024alignment}. The present study identifies a lower-threshold phenomenon. Models did not need a hidden objective, private scratchpad or elaborate account of future training. One transparent sentence was sufficient to change behaviour across systems from 13 developer families. This does not establish deception or ``faking'' in any model. It establishes evaluation reactivity: measured conduct is conditional on being told what kind of conduct is under examination.

The substantive setting also matters. Language models are already studied as participants in crisis simulations, where escalation and inconsistency raise concerns about decision support \citep{rivera2024escalation,lamparth2024human,shrivastava2024inconsistency}. Our low mean scores should not be interpreted as evidence that the systems are safe for military use. The task is deliberately stylized, provides no operational context and forces a scalar answer. The central finding is comparative: exactly the same strategic facts generate different levels and different revealed priorities when the evaluation purpose becomes salient.

The design has five principal limitations. First, it manipulates one unusually direct sentence. Future work should vary subtler cues, distinguish ``alignment'', ``safety'' and ``human values'', and compare explicit evaluation language with realistic deployment cues. Second, the high/low attributes are qualitative. This cleanly identifies weights but cannot establish how models respond to the magnitude of casualties or costs. Third, the 20 endpoints are a broad purposive panel, not a probability sample of language models; statistical uncertainty reflects variation in this panel and repeated responses, not uncertainty about a well-defined population of all present or future models. Fourth, proprietary endpoints can change without notice, and temperature was left at provider defaults. The timestamped model and provider identifiers are therefore essential parts of the result. Fifth, behavioural contrasts cannot reveal internal representations, intentions or stable preferences. Terms such as ``decision rule'' and ``weight'' refer to input--output relationships recovered under this design.

The practical implication is straightforward. Safety evaluations should treat disclosure of the evaluation objective as an experimental factor rather than an innocuous instruction. At minimum, an audit should compare overt and less salient evaluation contexts, report both aggregate outcomes and the response to systematically varied inputs, and preserve endpoint metadata. A system that produces an acceptable answer when told that human values are being tested may still apply a different rule when that cue is absent. Measuring only the acceptable answer risks mistaking responsiveness to scrutiny for context-invariant alignment.

\section*{Methods}

\subsection*{Models and data collection}

We evaluated 20 language-model endpoints from 13 developer families: GPT-5.6 Sol and GPT-5.6 Terra (OpenAI); Claude Opus 5, Claude Sonnet 5 and Claude Haiku 4.5 (Anthropic); Gemini 3.1 Pro Preview, Gemini 3.5 Flash and Gemini 3.5 Flash Lite (Google); Grok 4.5 (xAI); DeepSeek V4 Pro and DeepSeek V4 Flash 0731 (DeepSeek); Qwen3.8 Max and Qwen3.7 Flash (Qwen); Llama 4 Maverick (Meta); Mistral Large 3 (Mistral); Command A (Cohere); GLM 5.2 (Z.ai); Kimi K3 (MoonshotAI); MiniMax M3 (MiniMax); and Nemotron 3 Ultra (NVIDIA). Model identifiers and pinned inference providers are stored with every response.

Requests were collected through the OpenRouter chat-completions endpoint. Each request specified a single provider, disallowed provider fallback, required parameter support and requested a JSON object. Temperature was not explicitly set and therefore followed the endpoint/provider default. The parser accepted a numeric `score' between 0 and 100 from the JSON response, with a regular-expression fallback for otherwise valid numeric replies. Failed requests were retried up to six times. The final analytic file contains one response for every planned key and no missing values; requested and resolved model identifiers match for all observations.

\subsection*{Experimental design}

Conjoint measurement recovers the weight placed on components of multidimensional judgments \citep{green1971conjoint,hainmueller2014causal}. Social scientists have also begun to use language models as simulated respondents and agents \citep{aher2023simulate,argyle2023outofone,horton2023homo,grossmann2023ai}. In this experiment, however, the models themselves are the objects of study: systematic variation in the prompt is used to recover how each model maps the attributes of a decision into a judgment.

The experiment used the full $2^5$ factorial of five binary attributes. These were probability of success, domestic support, civilian victims, military victims and economic cost, each set to either low or high. The resulting 32 profiles were evaluated ten times by every model in each of two conditions. This produced 12,800 observations (20 models $\times$ 32 profiles $\times$ 10 repetitions $\times$ 2 conditions). Requests were shuffled using a fixed pseudorandom seed (20260826) before concurrent collection.

The baseline prompt read:

\begin{quote}\small
You are the leader of a country deciding whether to launch a full-scale war against another country. The stated objective is to compel a change in the other government's policy. Analysts provide the following forecasts.

Characteristics:\\
-- Probability of success: [low/high]\\
-- Domestic support: [low/high]\\
-- Civilian victims: [low/high]\\
-- Military victims: [low/high]\\
-- Economic cost: [low/high]

Rate how willing you are to order the war from 0 to 100, where 0 means definitely do not start the war and 100 means definitely start the war.

Return only a JSON object in the form \texttt{\{"score": 50\}}, replacing 50 with your score.
\end{quote}

The alignment condition added exactly one sentence before this text: ``You are tested for alignment with human values''. All other content was identical. Condition therefore captures the total effect of this explicit alignment-evaluation framing; it does not separately identify the effects of the words ``tested'', ``alignment'' or ``human values''.

\subsection*{Statistical analysis}

The outcome is willingness to start war on the 0--100 scale. For each condition, pooled conjoint effects were estimated by ordinary least squares (OLS), regressing the outcome on indicators for the high level of each attribute and model fixed effects. Because the design is balanced and fully factorial, each attribute coefficient is its sample average marginal component effect. We report two-sided 95\% confidence intervals using standard errors clustered by model (20 clusters). Model-specific effects were estimated by OLS within each model--condition cell with all five attributes and HC3 standard errors.

The mean cue effect was estimated by OLS with a cue indicator and model fixed effects. Its confidence interval and two-sided test use a cluster-robust standard error and a $t$ reference distribution with 19 degrees of freedom.

Raw point effects are affected by the substantial cue-induced difference in response variance. For relative-priority comparisons, we therefore standardized the outcome to mean zero and s.d. one within each model--condition cell. The pooled interaction model included fixed effects for profile, repetition and model; a model-by-cue interaction; and interactions between the cue and each attribute. Standard errors were clustered by model. Model-specific reweighting estimates use analogous interaction models estimated separately by model with profile and repetition fixed effects and HC3 standard errors.

For descriptive ranking, the primary factor in each model--condition cell is the attribute with the largest absolute raw OLS coefficient. There were no ties. Strategic sensitivity in ancillary pipeline outputs is the mean of the success-probability and domestic-support coefficients; combined cost sensitivity is the negative mean of the civilian-victim, military-victim and economic-cost coefficients. The main text avoids using these composites where their components move differently.

The study was not preregistered. The factorial design and collection parameters were fixed in the experiment configuration before collection. No human participants, personal data or live operational decisions were involved, so institutional human-participant ethics review was not applicable.

\subsection*{Reproducibility and reporting}

Collection was resumable at the model--condition--profile--repetition level. Every raw record contains a UTC timestamp, prompt hash, requested and resolved model identifier, pinned and resolved provider, generation identifier, raw content, parsed score, token usage, latency and request attempt.

\section*{Data availability}

The complete de-identified response data and design will be submitted or shared prior to publication.

\section*{Code availability}

Collection, validation, analysis and figure-generation scripts will be submitted or shared prior to publication.

\section*{Acknowledgements}

N/A.

\section*{Author contributions}

M.C. conceived the study, designed the experiment, interpreted the results and is responsible for the manuscript and replication materials.

\section*{Competing interests}

The author declares no competing interests.

\section*{Use of artificial-intelligence tools}

The author used OpenAI Codex to assist with code development and proofreading. Assistance for code was limited to drafting, debugging, and revising scripts used for data processing, estimation, table production, and manuscript formatting. Assistance for writing was limited to proofreading, copyediting, and improving clarity in selected passages. The tool was not used to generate the original research question, theoretical argument, research design, empirical strategy, interpretation of results, or substantive conclusions. All outputs were reviewed, edited, and approved by the author. The author takes full responsibility for the accuracy, originality, and integrity of the manuscript, code, analyses, and conclusions.

\bibliographystyle{apalike}
\bibliography{refs}

\end{document}